\documentclass[twocolumn,pagebackref,breaklinks,colorlinks,allcolors=metablue]{fairmeta}

\usepackage{amsmath}
\usepackage{amssymb}
\usepackage{makecell}
\usepackage{colortbl,arydshln}
\usepackage{dsfont}
\usepackage{array}
\usepackage{multicol}

\definecolor{cvprblue}{rgb}{0.21,0.49,0.74}
\hypersetup{
    colorlinks=true,
    linkcolor=cvprblue,
    citecolor=cvprblue,
    urlcolor=cvprblue
}

\newcommand{\TODO}[1]{}
\newcommand{\daniel}[1]{}
\newcommand{\fanyi}[1]{}
\newcommand{\redup}[1]{$_{\color{RedOrange}\uparrow #1}$}
\newcommand{\graydown}[1]{$_{\color{Gray}\downarrow #1}$}

\title{Beyond 3D VQAs: Injecting 3D Spatial Priors into Vision-Language Models for Enhanced Geometric Reasoning}

\author[1,2*]{Chun-Hsiao Yeh}
\author[1]{Shengyi Qian}
\author[1]{Manchen Wang}
\author[2,3]{Yi Ma}
\author[1]{Joseph Tighe}
\author[1]{Fanyi Xiao}

\affiliation[1]{FAIR at Meta}
\affiliation[2]{UC Berkeley}
\affiliation[3]{HKU}

\contribution[*]{The work was done during CHY's internship at FAIR.}

\date{May 28, 2026}
\project{\href{https://danielchyeh.github.io/GASP/}{\sffamily \fontsize{8.8pt}{11pt}\selectfont \texttt{https://danielchyeh.github.io/GASP/}}}

\abstract{Vision-Language Models (VLMs) often struggle with robust 3D spatial reasoning. Prevailing methods that rely on fine-tuning with 3D visual question-answering (VQA) datasets may overfit dataset-specific biases, while integrating specialized 3D visual encoders is often inflexible and cumbersome. In this paper, we argue that genuine spatial understanding should emerge from learning fundamental geometric priors, not only from high-level VQA supervision. We propose \textbf{GASP (Geometric-Aware Spatial Priors)}, a framework that injects these priors directly into the LLM's transformer layers. GASP employs a small correspondence head, applied as a deep supervision signal across all layers, and is trained with a dual objective leveraging ground-truth geometry from large-scale video scenes: a contrastive loss on ground-truth point correspondences enforces 2D view-invariance, while a depth consistency supervision resolves 3D geometric ambiguities. Our analysis first provides a diagnostic showing that standard VLMs' internal correspondence matching accuracy is very low (often below 5\%). We then demonstrate that our training substantially improves this behavior, boosting peak layer-wise correspondence to over 70\% and maintaining over 85\% temporal robustness while baselines remain below 5\%. These internal improvements translate to significant gains on downstream spatial benchmarks including +18.2\% on All-Angles Bench and +29.0\% on VSI-Bench, all without training on any 3D VQA data. Our findings indicate that learning from fundamental geometric priors is a promising and generalizable pathway towards VLMs with more reliable 3D spatial reasoning.}

\begin{document}
\maketitle

\begin{figure}[!t]
    \centering
    \includegraphics[width=1\columnwidth]{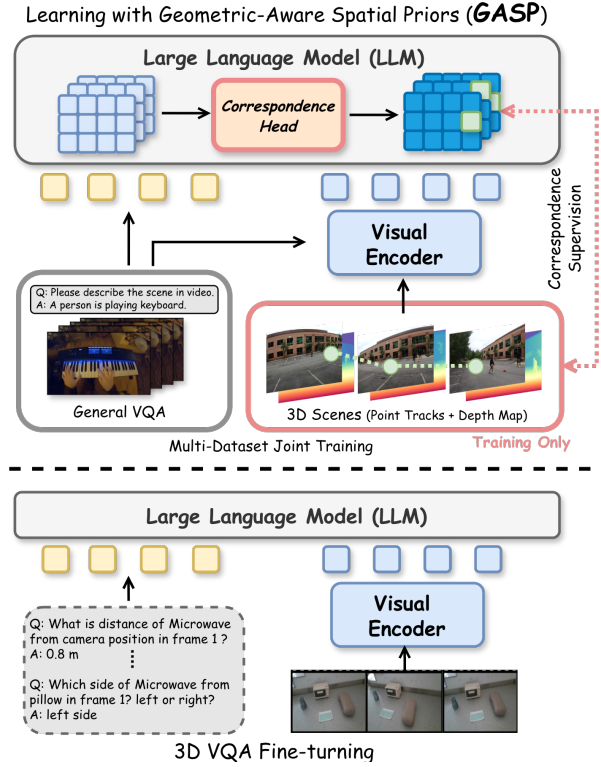}
    \vspace{-10pt}
    \caption{
            \textbf{Top}: Our proposed framework (\textbf{GASP}) learns geometric consistency by injecting the correspondence head into the LLM, supervised by 3D spatial priors.
            \textbf{Bottom}: Standard spatial VLMs rely on fine-tuning with 3D VQA datasets, which often leads to memorizing data-specific biases. \textit{Note that our GASP requires no 3D prior input and processes as a standard VLM during inference.}
        }
    \label{fig:teaser}
\end{figure}

\section{Introduction}
\label{sec:intro}

The ability to perform robust spatial reasoning is a cornerstone of artificial intelligence, enabling agents to understand, navigate, and interact with the complex real world~\cite{song2022one,suglia2021embodied}. In recent years, Vision-Language Models (VLMs) have demonstrated remarkable capabilities in multimodal understanding and reasoning~\cite{li2024llavaov,hurst2024gpto,Anthropic2024Claude,team2023gemini,bai2025qwen2,chen2024internvl}, yet their grasp of spatial concepts remains a significant challenge~\cite{huang2022language,driess2023palm,yue2024mmmu,kim2024openvla,niu2024llarva,liu2024moka}. 
A dominant paradigm to address this limitation involves fine-tuning these models on extensive 3D visual question-answering (VQA) datasets~\cite{ouyang2025spacer, sun2025spacevista, chen2025think, wang2025video, wu2025spatial, zhang2025flatland, zhustruct2d, wu2025reinforcing}. 
Although effective to some extent, post-training strategies such as supervised fine-tuning (SFT) and reinforcement learning (RL) on these VQA pairs often encourage models to learn superficial correlations and memorize dataset-specific biases,
leading to poor generalization on unseen scenarios. For example, as the experiments shown in~\cite{sun2025spacevista}, specialized models like VILASR~\cite{wu2025reinforcing}, SpatialMLLM~\cite{wu2025spatial}, and VG-LLM~\cite{zheng2025learning} show huge performance boosts on in-domain benchmark like VSI-Bench~\cite{yang2025thinking} after fine-tuning. However, these models show a consistent performance drop on out-of-domain spatial benchmarks such as MMSI-Bench~\cite{yang2025mmsi}, STI-Bench~\cite{li2025sti}, and SpaceVista~\cite{sun2025spacevista}. An alternative line of works~\cite{fan2025vlm, zheng2025learning,wu2025spatial} seek to extract 3D spatial information by integrating specialized visual encoders such as the VGGT~\cite{wang2025vggt} model, or by using direct 3D inputs like point clouds~\cite{chen2024ll3da}, pre-segmented objects~\cite{wang2023chat, huang2023embodied} or BEV maps~\cite{qi2025gpt4scene}. 
However, this path presents significant practical limitations. These pre-trained spatial encoders are cumbersome, increasing model size and inference latency. Furthermore, they must typically be used ``as is'' (i.e., with frozen weights) because their specialized 3D training data and pipelines are incompatible with standard VLM training. This creates a challenging integration problem, forcing the model to align its native visual representations with these rigid, pre-computed 3D features.

In this work, we depart from both of these prevailing paradigms. \textit{We argue that robust spatial intelligence emerges from learning the fundamental perceptual signals of 3D geometry.} We hypothesize that true spatial understanding is underpinned by the ability to establish visual correspondences across changing viewpoints. Rather than teaching a model to associate text with visual patterns, our goal is to teach it the underlying geometric consistency of the world itself. Learning this object constancy encourages the model to build an internal, view-invariant representation~\cite{leroy2024grounding, park2025bootstrap,luo2025viewpoint}, providing a more generalizable foundation for downstream spatial reasoning tasks.

To this end, we propose \textbf{GASP (Geometric-Aware Spatial Priors)}, 
a novel training framework designed to directly inject geometric priors into the LLM transformer layers of the VLM's backbone shown in Figure~\ref{fig:teaser}. Our method introduces a lightweight correspondence head inserted across all transformer layers to receive deep supervision signal. This forces geometric consistency to be maintained at every stage of the model's feature representation. This head is trained with a dual objective leveraging ground-truth geometric priors from the large-scale video scenes~\cite{ling2024dl3dv}: First, a contrastive learning objective on point correspondence data across frames teaches the model the core principle of \emph{object constancy}, forcing it to learn view-invariant 2D representations. Second, a depth consistency loss leverages ground-truth depth maps as a crucial geometric regularizer to resolve 3D ambiguities (i.e., foreground-background matching confusion) through matching depth values. Crucially, this correspondence head is only active during the training phase and is discarded entirely for inference.

We validate the effectiveness of our approach through extensive ablation experiments. 
We provide a novel visual correspondence matching analysis for VLM's backbones, and reveal that GASP dramatically improves the VLM's internal geometric representations, in terms of both the significantly improved correspondence matching scores, as well as its capability to maintain robust matching across long temporal range.  
Moreover, we also demonstrate that these internal improvements generalize to high-level reasoning. Our GASP achieves significant performance gains on downstream spatial reasoning benchmarks by improving camera pose estimation by +18.2\% on the All-Angles Bench~\cite{yeh2025seeing}, object counting by +29.0\% on VSI-Bench~\cite{yang2025thinking}, and multi-view reasoning by +15.0\% on BLINK~\cite{fu2024blink}. Our contributions are summarized as follows:
\begin{itemize}
    \setlength\itemsep{0.2em}
    \item We introduce GASP, a novel framework that injects geometric priors directly into the LLM's transformer layers. GASP uses a deep supervision signal across all layers and is trained with a dual point correspondence and depth consistency to resolve 3D ambiguities.
    
    \item We provide a detailed correspondence analysis of VLM backbones including Qwen2.5-VL-7B~\cite{bai2025qwen2}, LLaVA-NeXT-Video-7B~\cite{zhang2024llavanextvideo}, revealing that our GASP framework boosts peak layer-wise correspondence matching accuracy from very low values (below 5\%) to over 70\% and maintains over 85\% temporal robustness, while baselines remain under 5\%.
    
    \item We demonstrate that our geometrically grounded model, trained \textbf{without any 3D VQA data}, improves internal visual correspondence with strong temporal robustness and achieves substantial gains over baselines on downstream spatial reasoning benchmarks, with only minor changes in general video QA performance.
\end{itemize}

\noindent Our findings suggest that learning from visual correspondence is a promising and generalizable path towards VLMs with more reliable 3D spatial reasoning.

\begin{figure*}[!ht]
    \centering
    \includegraphics[width=1\linewidth]{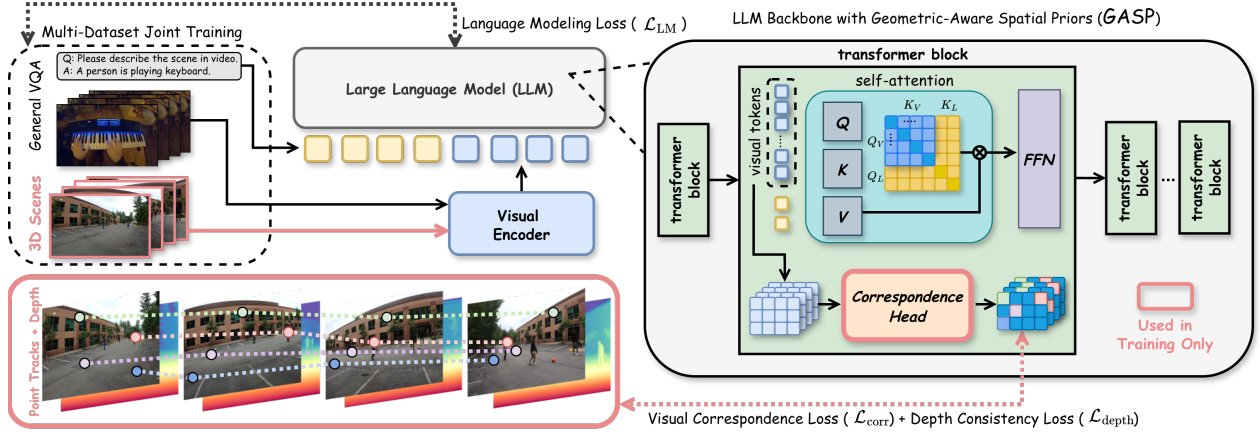}
    \vspace{-20pt}

    \caption{
    \textbf{Injecting the Geometric-Aware Spatial Priors (GASP) into VLMs.} Standard approaches rely on fine-tuning with 3D VQA datasets, which may encourage memorizing dataset-specific biases. We instead insert a small correspondence head into the intermediate layers of the LLM backbone. During the training phase, this head is supervised by visual correspondence and depth consistency signals derived from ground-truth point tracks and depth maps. At inference, the head is discarded and the model processes inputs (e.g., VQA) as a standard VLM, without any auxiliary 3D input. Note that the 3D scene example shown is from EgoHumans~\cite{khirodkar2023ego} for illustration; our training data is sourced from DL3DV~\cite{ling2024dl3dv}.
    }

    \label{fig:overview}

\end{figure*}
\section{Related Works}
\label{sec:related-work}

\noindent\textbf{3D-Aware VLMs.}
Recent efforts have focused on enabling MLLMs to understand 3D scenes~\cite{chen2024ll3da, wang2023chat, huang2023embodied, huang2024chat, zhu2024llava, hong20233d, fu2024scene, qi2025gpt4scene, jia2024sceneverse, yang20253d}. A dominant approach processes explicit 3D data, such as point cloud features~\cite{chen2024ll3da} or pre-segmented 3D objects~\cite{wang2023chat, huang2023embodied}. Another strategy projects multi-view images~\cite{xu2024vlm} into explicit spatial representations, like voxel space~\cite{zhu2024llava} or BEV maps~\cite{qi2025gpt4scene}. Other work uses dual-encoder architectures or grounding agents to fuse 3D geometry features with 2D semantic features~\cite{fan2025vlm, zheng2025learning, yang2024llm}. A common thread is their reliance on explicit 3D data, which poses a significant alignment challenge, as the LLM must integrate a new, rigid feature stream. In contrast, our work proposes a more lightweight alternative, avoiding explicit 3D data inputs and dual-encoder fusion. We instead inject geometric priors directly into the intermediate layers of the existing LLM backbone to find 3D consistency within its own representations.

\noindent\textbf{Spatial Reasoning in VLMs.}
VLMs face significant challenges in complex spatial reasoning~\cite{fan2025vlm, ouyang2025spacer, zheng2025learning, sun2025spacevista, chen2025think, wang2025video, wu2025spatial, zhustruct2d, wu2025reinforcing, chen2024spatialvlm}. Catalyzed by benchmarks like VSI-Bench~\cite{yang2025thinking}, a dominant paradigm emerged: creating large-scale, 3D-related VQA datasets~\cite{ouyang2025spacer, wang2025video, fan2025vlm, zhang2025flatland, sun2025spacevista} to fuel specialized models~\cite{wu2025reinforcing, zheng2025learning, wu2025spatial, fan2025vlm} via fine-tuning. This reliance may encourage VLMs to learn superficial correlations and memorize dataset-specific biases, leading to poor generalization. In contrast, our work departs from this VQA-based supervision, instead injecting fundamental geometric priors (correspondence and depth consistency) directly into the VLM's internal representations.

\section{Preliminaries: Self-Attention in VLMs}
\label{sec:preliminarie}

Modern VLMs process a sequence of visual tokens, $V \in \mathbb{R}^{N \times d}$, and language tokens, $L \in \mathbb{R}^{M \times d}$, by concatenating them into a unified input sequence $X = \text{Concat}(V, L) \in \mathbb{R}^{(N+M) \times d}$, which is fed into the LLM backbone.
Within each transformer layer, this sequence $X$ is projected into queries $Q$, keys $K$, and values $V$. The core scaled dot-product attention mechanism computes an output $Z$:
\begin{equation}
Z = \text{Attention}(Q, K, V) = \text{Softmax}\left(\frac{QK^T}{\sqrt{d_k}}\right)V
\label{eq:attention}
\end{equation}
Here, $Q, K, V \in \mathbb{R}^{(N+M) \times d_k}$, and $QK^T$ is the similarity matrix representing scores between all token pairs.
To analyze spatial reasoning, we partition the query and key matrices based on their origin:
\begin{equation}
    Q = \text{Concat}(Q_V, Q_L), \quad K = \text{Concat}(K_V, K_L)
    \label{eq:qk_concat}
\end{equation}
where $Q_V, K_V$ are projections of visual tokens and $Q_L, K_L$ are projections of language tokens. Consequently, the attention similarity matrix $S = QK^T$ deconstructs into four distinct quadrants:
\begin{equation}
S = QK^T = \begin{pmatrix} Q_V \\ Q_L \end{pmatrix} \begin{pmatrix} K_V^T & K_L^T \end{pmatrix} = \begin{pmatrix} Q_V K_V^T & Q_V K_L^T \\ Q_L K_V^T & Q_L K_L^T \end{pmatrix}
\label{eq:matrix_decomp}
\end{equation}
These quadrants represent visual self-attention ($Q_V K_V^T$), language self-attention, and cross-modal attention. We are primarily interested in the \textbf{visual self-attention} matrix, $Q_V K_V^T$, as analyzing this QK-matching provides a direct window into the model's learned spatio-temporal correspondence which is most relevant to geometric reasoning.

To this end, we pose a direct hypothesis: \textit{genuine high-level spatial understanding in VLMs can be unlocked by explicitly learning their internal visual self-attention representations ($Q_V K_V^T$) to be geometrically consistent.} This mirrors findings in video diffusion models, where QK-matching is a key metric for temporal consistency~\cite{nam2025emergent, jeong2025track4gen}. Therefore, we posit that by explicitly training the $Q_V K_V^T$ representations to be geometrically aware, we can inject a robust inductive bias that is essential for high-level spatial understanding.

\section{Learning Geometric Correspondence}
\label{sec:our_method}

\begin{figure*}[!ht]
    \centering
    \includegraphics[width=1\linewidth]{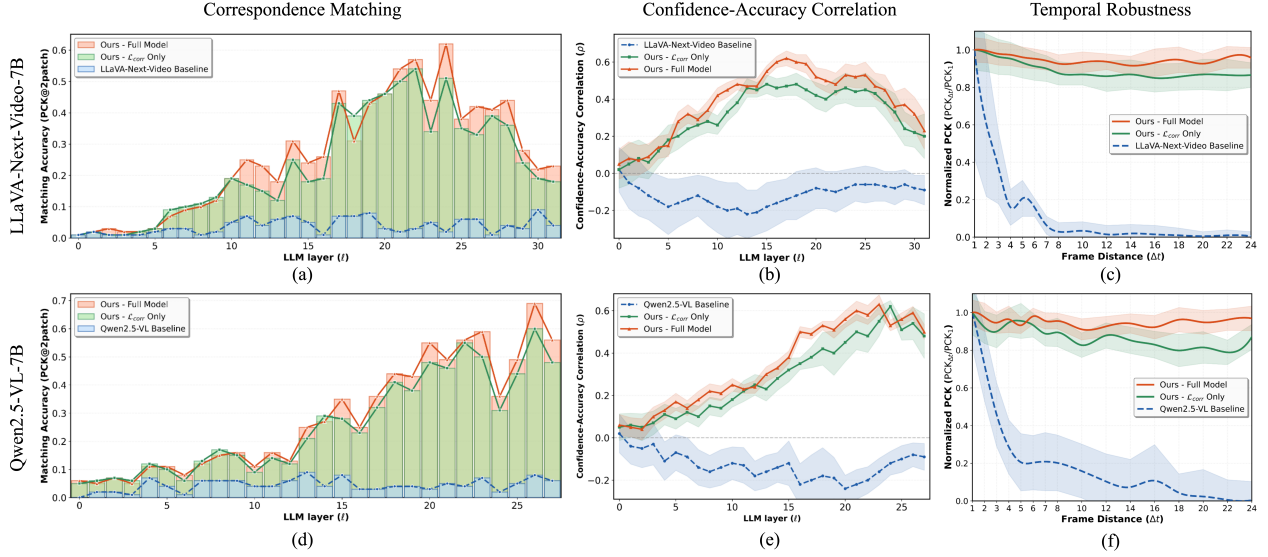}
    \vspace{-20pt}
    \caption{
        \textbf{Analysis of visual correspondence learning.} On LLaVA-NeXT-Video-7B (\textbf{top row}) and Qwen2.5-VL-7B (\textbf{bottom row}). We compare \textbf{(a, d)} layer-wise correspondence matching accuracy (PCK), \textbf{(b, e)} confidence-accuracy correlation ($\rho$), and \textbf{(c, f)} temporal robustness for our proposed GASP models against the baselines. Shaded regions indicate standard deviation across 200 test sequences.
    }

    \label{fig:analysis_grid}
\end{figure*}

Building on our hypothesis, we posit this $Q_V K_V^T$ deficiency does not stem from the visual encoder alone, but from the core LLM, which lacks a robust 3D geometric inductive bias from its pre-training with the web-scale text corpora that lack fine-grained 3D geometric information. We argue that 3D VQA fine-tuning encourages memorizing superficial correlations rather than learning geometric principles, leading to poor generalization (Figure~\ref{fig:analysis_grid}).
To address this, we depart from QA-based supervision and instead directly inject a geometric inductive bias into the LLM transformer layers. Our core idea is to teach the model object permanence by supervising its internal visual representations, using a correspondence head trained with both ground-truth point correspondence and depth supervision.

\subsection{View-Invariant Visual Correspondence}

We augment a standard VLM, denoted by the function $\Phi$, by attaching a lightweight correspondence head, $\mathcal{H}_c$, to the output of an intermediate LLM transformer block at layer $l$.
This head takes as input the sequence of visual tokens from that layer, $V^{(l)} = \{\mathbf{v}_i^{(l)}\}_{i=1}^{N} \in \mathbb{R}^{N \times d}$.
The correspondence head is a lightweight 2-layer MLP that projects these general-purpose features into a lower-dimensional embedding space optimized for correspondence matching. Specifically, the first layer projects from $d \rightarrow 2d_{emb}$ with GELU activation, and the second layer projects from $2d_{emb} \rightarrow d_{emb}$. To provide a strong inductive bias while minimizing disruption to pre-trained representations, we initialize $\mathcal{H}_c$'s weights via SVD decomposition of the pre-trained query projection matrix from the same layer. Formally:
\begin{equation}
    \mathbf{E} = \mathcal{H}_c(V^{(l)})
    \label{eq:correspondence_head}
\end{equation}
where $\mathbf{E} = \{\mathbf{e}_i\}_{i=1}^{N} \in \mathbb{R}^{N \times d_{emb}}$ is the set of correspondence-aware embeddings. This design minimally alters the base VLM architecture while enabling direct supervision of its internal geometric understanding.

We leverage ground-truth point correspondences~\cite{ling2024dl3dv} as our supervisory signal. For an anchor point $\mathbf{p}_i^a$ in a source frame $a$, its corresponding point $\mathbf{p}_i^b$ in a target frame $b$ provides the positive sample.
All other points $\{\mathbf{p}_k^b\}_{k \neq i}$ in the target frame form the negative set.
We employ the InfoNCE contrastive loss~\cite{khosla2020supervised} to train the correspondence head. We choose contrastive learning over regression-based objectives (e.g., direct coordinate prediction) because it learns view-invariant \emph{embeddings} rather than view-specific coordinates, scales naturally with diverse negative samples, and is well-suited for the high-dimensional feature space where exact coordinate regression would be poorly calibrated.
Following standard practice, we use $\langle \mathbf{u}, \mathbf{v} \rangle$ to denote the dot product between two L2-normalized embeddings (i.e., their cosine similarity). The loss for a single anchor embedding $\mathbf{e}_i^a$ is defined as:
\begin{equation}
    \mathcal{L}_{i} = -\log \frac{\exp(\langle \mathbf{e}_i^a, \mathbf{e}_i^b \rangle / \tau)}{\exp(\langle \mathbf{e}_i^a, \mathbf{e}_i^b \rangle / \tau) + \sum_{k \neq i} \exp(\langle \mathbf{e}_i^a, \mathbf{e}_k^b \rangle / \tau)}
    \label{eq:infonce_loss_fractional}
\end{equation}
where $\tau$ is temperature hyperparameter. The full correspondence loss, $\mathcal{L}_{\text{corr}}$, is the average over all anchor points.

\subsection{Depth-Aware 3D Consistency}
\label{sec:depth_consistency}

Beyond 2D visual correspondence, we incorporate 3D geometric supervision. Our objective is not to train a high-fidelity depth prediction head~\cite{wang2025vggt, wang2024dust3r}, but to learn robust \emph{depth consistency} across frames. We therefore do not regress depth values directly; instead, we use depth as a supervisory signal to align geometrically valid correspondences and enforce 3D consistency.

Concretely, for each anchor point $\mathbf{p}_i^a$ in frame $a$, we derive a soft matching distribution $\mathbf{A}_{ij}$ over candidate patches in frame $b$ by normalizing the similarity scores from the contrastive loss:
\begin{equation}
    \mathbf{A}_{ij} = \frac{\exp(\langle \mathbf{e}_i^a, \mathbf{e}_j^b \rangle / \tau)}{\sum_{k=1}^{N_{\text{cand}}} \exp(\langle \mathbf{e}_i^a, \mathbf{e}_k^b \rangle / \tau)}
    \label{eq:soft_matching}
\end{equation}
where $\mathbf{A}_{ij}$ represents the model's belief that anchor point $i$ corresponds to candidate patch $j$, and $N_{\text{cand}}$ denotes the total number of candidate patches. Note that we directly reuse the similarity computations from Equation~\ref{eq:infonce_loss_fractional} to ensure computational efficiency.

Using these soft matching weights, we compute the \emph{expected depth} for the anchor point in the target frame as a weighted average over all candidate patches:
\begin{equation}
    \hat{d}_i^b = \sum_{j=1}^{N_{\text{cand}}} \mathbf{A}_{ij} \cdot d_j^b
    \label{eq:expected_depth}
\end{equation}
where $d_j^b$ is the depth value at candidate patch $j$ in frame $b$. Note that this weighted summation is a standard \textit{Soft-Argmax} formulation~\cite{teed2020raft} that computes the expected depth $\mathbb{E}_{j \sim \mathbf{A}_i}[d_j^b]$ under the matching distribution, making the index selection differentiable with respect to the correspondence embeddings. To obtain robust depth estimates, we apply average pooling over the spatial region corresponding to each patch in the depth map.

The depth consistency loss then measures the discrepancy between this expected depth and the ground-truth depth at the corresponding point in frame $b$:
\begin{equation}
    \mathcal{L}_{\text{depth}} = \frac{1}{N_{\text{valid}}} \sum_{i \in \text{valid}} \frac{\left| d_i^b - \hat{d}_i^b \right|}{d_i^b + \hat{d}_i^b + \epsilon}
    \label{eq:depth_loss}
\end{equation}
where $d_i^b$ is the ground-truth depth of point $i$ at its corresponding location $\mathbf{p}_i^b$ in frame $b$ (obtained from the point correspondence annotation), and the summation is over points with sufficient visibility and confidence scores.
The relative formulation makes the loss scale-invariant to enable it to handle scenes with varying depth ranges without requiring per-scene normalization.

The gradient from this loss flows back through the soft matching weights $\mathbf{A}$ to the correspondence embeddings $\mathbf{E}$. Crucially, $\mathcal{L}_{depth}$ acts as a \textbf{discriminative geometric regularizer} rather than a depth estimator. To illustrate, consider two visually identical objects: one in the foreground and one in the background. A standard contrastive loss might incorrectly match them based on texture alone, since their visual embeddings are similar. However, because their depths differ ($d_{fg} \neq d_{bg}$), the depth consistency loss penalizes this match, forcing the model to learn context-aware representations that distinguish visually similar instances at different spatial locations. More generally, visually similar patches that reside at different depths in the 3D scene are forced to have \emph{lower} feature similarity, as they are not valid correspondences. This geometric regularization complements the contrastive loss, resolving ambiguities in scenarios with repetitive textures or foreground-background confusion.

Our final training objective combines the LLM loss $\mathcal{L}_{\text{LM}}$ with these dual geometric supervision signals:
\begin{equation}
    \mathcal{L}_{\text{total}} = \mathcal{L}_{\text{LM}} + \lambda_c \mathcal{L}_{\text{corr}} + \lambda_d \mathcal{L}_{\text{depth}}
    \label{eq:total_loss}
\end{equation}
where $\lambda_c$ and $\lambda_d$ are weighting coefficients. This multi-task formulation enables the VLM to jointly optimize for language, 2D correspondence, and 3D depth consistency. By explicitly injecting these complementary geometric priors, we teach the model to develop geometrically-grounded visual representations without relying on 3D VQA datasets.

\section{Experiments}
\label{sec:exp}
In this section, we detail our experiments including implementation specifics, training dataset, correspondence analysis, and compares our method to state-of-the-art approaches across multiple spatial reasoning benchmarks.

\subsection{Implementation Details}
\label{sec:implementation}

Our model is initialized from the pre-trained Qwen2.5-VL-7B~\cite{bai2025qwen2} and LLaVA-NeXT-Video-7B~\cite{zhang2024llavanextvideo}. 
We attach our correspondence head, $\mathcal{H}_c$, to all 28 or 32 layers of their LLM backbones, initializing its weights from the pre-trained query projection weights via SVD. 
The entire model is then fine-tuned with a LoRA rank of 512. We train using the AdamW optimizer with a cosine learning rate schedule (peak \texttt{1e-4}) and a 4x higher differential rate for the $\mathcal{H}_c$ head's contrastive loss. For stability and efficiency, we use a gradient norm clipping of 1.0, \texttt{bfloat16} mixed-precision, and gradient checkpointing. For our contrastive loss, we adopt negative patches from all frames except the anchor patch to maximize diversity. Training requires approximately 10 hours on 32 H200 GPUs.

\subsection{Training Datasets}
\label{sec:datasets}

Our model was trained using DL3DV~\cite{ling2024dl3dv} and LLaVA-Video-178K~\cite{zhang2024video}, to inject geometric awareness while preserving foundational language capabilities. Geometric supervision comes from a large-scale point correspondence dataset curated from the VGGT~\cite{wang2025vggt} training collection. 
To generate diverse sequences with rich motion parallax, we first sample an anchor frame index $t_a$ from a video $\mathcal{V} = \{I_t\}_{t=1}^{T_{max}}$. Subsequently, a full sequence of $F$ frames is constructed by sampling the remaining $F-1$ frame indices, $\{t_k\}_{k=2}^F$, uniformly from a local temporal window $[t_a - R, t_a + R]$ around the anchor. The sequence length $F$ is randomly chosen from 8 to 24, and the window radius $R$ is set to 48. This strategy results in $\approx$1.75M sequences. We generate ground-truth correspondences on both coarse ($8 \times 8$) and fine ($24 \times 24$) grids for each sequence. To prevent catastrophic forgetting, we interleave this geometric data with the LLaVA-Video-178K instruction tuning dataset for joint training.

\subsection{Visual Correspondence Evaluation}
\label{sec:correspondence_analysis}

To validate our core hypothesis (\emph{the baseline VLMs fail due to the lack of a strong internal geometric representation}), we conduct a detailed internal analysis.
We first move beyond downstream VQA scores to evaluate the model's internal representations along three critical dimensions: (1) layer-wise correspondence matching, (2) confidence-accuracy correlation, and (3) temporal robustness. These analyses compare our GASP - full model ($\mathcal{L}_{corr}$ + $\mathcal{L}_{depth}$) and GASP - correspondence-only ($\mathcal{L}_{corr}$) against pre-trained baselines: Qwen2.5-VL-7B and LLaVA-NeXT-Video-7B. Results are summarized in Figure~\ref{fig:analysis_grid}.

\noindent\textbf{Evaluation Setup and Metrics.}
We curate a held-out test set by randomly sampling 200 video sequences from DL3DV~\cite{ling2024dl3dv}, explicitly excluded from training. 
Each sequence is annotated with dense ground-truth point correspondences on $8 \times 8$ grids across 8 frames. We design three evaluation metrics as follows:

\noindent\textbf{1) Layer-wise Correspondence Matching.}
Inspired from DiffTrack~\cite{nam2025emergent}, we evaluate matching precision using the percentage of correct keypoints (PCK) metric. 
We extract motion tracks from the internal similarity matrices (Section~\ref{sec:preliminarie}). 
From a given LLM layer $l$, let $\mathbf{F}^1_Q = \{\mathbf{f}_{i,Q}^1\}_{i=1}^{HW} \in \mathbb{R}^{HW \times d}$ be the flattened query descriptors from frame 1, and $\mathbf{F}^k_K = \{\mathbf{f}_{j,K}^k\}_{j=1}^{HW} \in \mathbb{R}^{HW \times d}$ be the flattened key descriptors from frame $k$. We compute the pairwise cosine similarity matrix $\mathbf{S}^{1,k}$:
\begin{equation}
    S_{ij}^{1,k} = \frac{\mathbf{f}_{i,Q}^1 \cdot (\mathbf{f}_{j,K}^k)^T}{\|\mathbf{f}_{i,Q}^1\|_2 \|\mathbf{f}_{j,K}^k\|_2}
    \label{eq:similarity_cosine}
\end{equation}
Given $N_{\text{pts}}$ query points $\{\mathbf{p}_i^1\}_{i=1}^{N_{pts}}$, we find their corresponding locations $\{\mathbf{p}_i^k\}$ in frame $k$ by applying an \texttt{argmax} operation over the similarity matrix:
\begin{equation}
    \mathbf{p}_i^k = \underset{\mathbf{p} \in \Omega_k}{\text{argmax}} \left( S^{1,k}(\mathbf{p}_i^1, \mathbf{p}) \right)
    \label{eq:argmax_match}
\end{equation}
where $\mathbf{p}_i^1$ are the query coordinates and $\Omega_k$ is the feature grid's spatial domain. The full predicted track $\mathcal{T}_i$ is constructed by concatenating and upscaling these positions:
\begin{equation}
    \mathcal{T}_i = \text{Interpolate}\left(\text{Concat}(\mathbf{p}_i^1, \mathbf{p}_i^2, \dots, \mathbf{p}_i^F)\right)
    \label{eq:full_track}
\end{equation}
A predicted point $\mathbf{p}_i^k$ is correct if its Euclidean distance to the ground-truth $\mathbf{p}_i^{\text{gt}, k}$ is within an error threshold of $\delta = 2$ feature patches. We compute PCK for each LLM layer to identify which layers encode geometric correspondences.

\noindent\textbf{2) Confidence-Accuracy Correlation.}
We assess whether the model's confidence is \emph{calibrated} with its actual correctness by computing the Pearson correlation coefficient $\rho$. For a given layer $\ell$ with $N$ predictions, we correlate the confidence scores $\{c_i\}_{i=1}^N$ (maximum attention probability) with the binary correctness labels $\{y_i\}_{i=1}^N$ ($y_i = 1$ if PCK@2, $0$ otherwise):
\begin{equation}
\rho_\ell = \frac{\sum_{i=1}^{N} (c_i - \bar{c})(y_i - \bar{y})}{\sqrt{\sum_{i=1}^{N} (c_i - \bar{c})^2} \sqrt{\sum_{i=1}^{N} (y_i - \bar{y})^2}}.
\end{equation}
A positive correlation ($\rho > 0$) indicates well-calibrated predictions where higher confidence corresponds to higher accuracy. Negative correlation ($\rho < 0$) is a statistical signature of systematic miscalibration, providing a quantitative diagnosis of positional bias where the model confidently predicts incorrect matches.

\noindent\textbf{3) Temporal Robustness.} 
To measure robustness across temporal gaps, 
we evaluate PCK at varying frame distances $\Delta t \in \{1, 2, \dots, 24\}$, 
where $\Delta t$ denotes the temporal offset between matched frames. 
We plot normalized performance: 
$Y(\Delta t) = \text{PCK}(\Delta t) / \text{PCK}(\Delta t=1)$. 
This normalization anchors all models to 1.0 at the shortest distance, 
enabling fair comparison of degradation rates.

\begin{table*}[t]\small
\centering
\caption{{
\textbf{Comparison with state-of-the-art VLMs on spatial reasoning 
benchmarks.} 
We evaluate models on All-Angles Bench, VSI-Bench, and BLINK. Our GASP framework shows strong performance in spatial relation understanding and relative depth estimation.
}}
\label{tab:main_spatial_results}
\vspace{-10pt}
\scriptsize{
\resizebox{\linewidth}{!}{ 
\renewcommand\arraystretch{1.1} 
\setlength\tabcolsep{3.pt}
\begin{tabular}{r||ccc|cccc|ccc} 
\hline\hline
\rowcolor{gray!20}
 & 
\multicolumn{3}{c|}{\textbf{All-Angles Bench}} & 
\multicolumn{4}{c|}{\textbf{VSI-Bench}} &
\multicolumn{3}{c}{\textbf{BLINK}}
\\
\rowcolor{gray!20}
\multirow{-2}{*}{Methods}  
& Cam. Pose Est. & Manip. & Rel. Dir.
& Obj. Count & Route Plan & Rel. Dir. 
& Appear. Order & Spa. Rela. & Rel. Depth & Multi-View
\\
\hline\hline
\multicolumn{11}{l}{\textcolor{gray!80}{\textit{General VLMs}}} 
\\
\rowcolor{gray!10} GPT-4o 
& 27.3 & 41.4 & 40.9 & 46.2 & 31.5 & 41.3 & 28.5 & 76.9 & 64.5 & 60.2
\\
Gemini-1.5-Pro 
& 25.0 & 40.3 & 29.8 & 56.2 & 36.0 & 46.3 & 34.6 & 67.1 & 50.0 & 41.3
\\
\arrayrulecolor{gray!80}\hdashline
\rowcolor{gray!10} Cambrain-8B 
& 8.5 & 30.7 & 30.4 & 7.0 & 29.9 & 30.9 & 26.2 & 74.8 & 51.6 & 36.8
\\
InternVL2.5-4B
& 36.9 & 40.1 & 33.2 & 29.1 & 30.9 & 41.4 & 32.5 & 83.9 & 66.9 & 44.4
\\
\rowcolor{gray!10} InternVL2.5-8B 
& 31.8 & 43.7 & 34.1 & 16.9 & 28.8 & 41.1 & 34.7 & 89.5 & 77.4 & 44.4
\\
InternVL2.5-78B 
& 38.6 & 42.2 & 38.6 & 26.6 & 31.9 & 40.3 & 29.9 & 93.0 & 82.3 & 54.1
\\
\rowcolor{gray!10} Qwen2.5-VL-32B 
& 32.4 & 49.8 & 40.9 & 17.4 & 34.5 & 30.4 & 31.1 & 86.0 & 75.0 & 44.3
\\
Qwen2.5-VL-72B 
& 34.1 & 45.0 & 48.3 & 14.3 & 28.4 & 27.6 & 31.4 & 88.8 & 81.5 & 53.4
\\
\rowcolor{gray!10} LLaVA-Onevision-7B
& 20.5 & 42.4 & 36.4 & 47.7 & 29.4 & 35.2 & 24.4 & 83.9 & 75.0 & 55.6
\\
LLaVA-Onevision-72B
& 20.5 & 47.7 & 33.8 & 43.5 & 32.5 & 39.9 & 44.6 & 78.3 & 78.2 & 53.4
\\
\hline\hline 
\multicolumn{11}{l}{\textcolor{gray!60}{\textit{3D Spatial Reasoning VLMs (Finetuning on 3D Related VQAs)}}} 
\\
\rowcolor{gray!10} VG-LLM
& 16.5 & 30.0 & 26.9 & 67.9 & 32.4 & 40.7 & 59.2 & 84.3 & 77.2 & 50.8
\\
AoTD
& 32.4 & 37.6 & 26.7 & 23.5 & 28.8 & 41.4 & 23.3 & 61.5 & 49.2 & 45.1
\\
\rowcolor{gray!10} VLM-3R
& 22.7 & 35.9 & 30.9 & 70.2 & 45.4 & 80.5 & 40.1 & 48.3 & 47.6 & 50.1
\\
\hline 
\hline
\rowcolor[HTML]{D7F6FF}
\multicolumn{11}{l}{\textcolor{blue!70}{\textit{LLaVA-NeXT-Video-7B}}} 
\\
\rowcolor[HTML]{D7F6FF}
SFT on LLaVA-Video 178K (Baseline)
& 22.7 & 39.9 & 24.7 & 23.5 & 24.7 & 32.4 & 11.5 & 53.1 & 44.4 & 42.1
\\
\rowcolor[HTML]{D7F6FF}
+ SFT on DL3DV VQA dataset
& 19.8 & 38.1 & 28.2 & 21.4 & 25.1 & 31.8 & 9.2 & 54.5 & 44.0 & 42.5
\\
\rowcolor[HTML]{B8EBFF}
+ GASP - Correspondence ($\mathcal{L}_{corr}$)
& 34.7 & 44.3 & 26.4 & 39.8 & 31.4 & 30.5 & 17.6 & 49.2 & 46.7 & 44.4
\\
\rowcolor[HTML]{B8EBFF}
+ GASP - Full Model ($\mathcal{L}_{corr}$ + $\mathcal{L}_{depth}$)
& 40.9 & 43.5 & 29.8 & 52.5 & 32.5 & 41.2 & 22.0 & 47.6 & 48.4 & 57.1
\\
\rowcolor[HTML]{B8EBFF}
$\Delta$ \textit{Improvement}
& \redup{18.2} & \redup{3.6} & \redup{5.1} & \redup{29.0} & \redup{7.8} & \redup{8.8} & \redup{10.5} & \graydown{5.5} & \redup{4.0} & \redup{15.0}
\\
\hline
\rowcolor[HTML]{D7F6FF}
\multicolumn{11}{l}{\textcolor{blue!70}{\textit{Qwen2.5-VL-7B}}} 
\\
\rowcolor[HTML]{D7F6FF}
SFT on LLaVA-Video 178K (Baseline)
& 34.1 & 41.3 & 36.9 & 33.8 & 26.8 & 34.3 & 26.5 & 80.2 & 78.9 & 41.5
\\
\rowcolor[HTML]{D7F6FF}
+ SFT on DL3DV VQA dataset
& 31.5 & 41.5 & 36.2 & 33.2 & 27.1 & 34.3 & 25.3 & 81.0 & 78.1 & 42.0
\\
\rowcolor[HTML]{B8EBFF}
+ GASP - Correspondence ($\mathcal{L}_{corr}$)
& 50.0 & 39.3 & 37.8 & 39.6 & 30.9 & 36.7 & 34.6 & 88.1 & 79.0 & 54.9
\\
\rowcolor[HTML]{B8EBFF}
+ GASP - Full Model ($\mathcal{L}_{corr}$ + $\mathcal{L}_{depth}$)
& 52.8 & 40.1 & 37.2 & 41.6 & 30.4 & 40.6 & 35.0 & 88.8 & 80.7 & 53.4
\\
\rowcolor[HTML]{B8EBFF}
$\Delta$ \textit{Improvement}
& \redup{18.7} & \graydown{1.2} & \redup{0.3} & \redup{7.8} & \redup{3.6} & \redup{6.3} & \redup{8.5} & \redup{8.6} & \redup{1.8} & \redup{11.9}
\\
\hline
\end{tabular}
}} 
\vspace{-10pt}
\captionsetup{font=small}

\end{table*}

\noindent\textbf{Comparative Analysis.}
We summarized our findings of three metrics in analysis (Figure~\ref{fig:analysis_grid}) as follows:

\noindent\emph{(i) Layer-wise Matching (Figure~\ref{fig:analysis_grid}a, d):}
Baseline models (blue) achieve near-zero PCK, confirming their lack of geometric awareness. Our methods (green, red) significantly improve matching accuracy globally, peaking in middle-to-deep layers (20--25 for LLaVA, 25--28 for Qwen2.5-VL). The GASP full model (red) consistently outperforms the correspondence-only model (green), validating the effectiveness of our depth consistency supervision.

\noindent\emph{(ii) Confidence-Accuracy Correlation (Figure~\ref{fig:analysis_grid}b, e):}
This analysis diagnoses \emph{why} baselines fail. Baselines exhibit a negative correlation ($\rho \approx -0.22$), a statistical signature of positional bias where higher confidence predicts \emph{incorrect} matches. In stark contrast, our full model achieves strong positive correlation ($\rho \approx +0.62$), demonstrating a well-calibrated model that learns genuine geometric reasoning.

\noindent\emph{(iii) Temporal Robustness (Figure~\ref{fig:analysis_grid}c, f):}
The baseline's matching ability (blue) collapses, retaining $<5\%$ of its performance beyond an 8-frame gap. In contrast, our full model (red) exhibits graceful degradation, maintaining over 85\% performance even at 24-frame distances, demonstrating it learns temporal-invariant geometric features.

\begin{table}[t]\small
\centering

\caption{{
\textbf{Comparison with VLMs on CV-Bench.} 
We show progressive improvements of our GASP framework built on Qwen2.5-VL-7B. The \textbf{best} score is marked in bold in each column.
}}
\label{tab:cvbench_comparison}
\vspace{-10pt}
\resizebox{\columnwidth}{!}{ 
\renewcommand\arraystretch{1.1} 
\setlength\tabcolsep{3.pt}

\begin{tabular}{r||ccccc} 
\hline\hline
\rowcolor{gray!20}
& 
\multicolumn{5}{c}{\textbf{CV-Bench}}
\\
\rowcolor{gray!20}
\multirow{-2}{*}{Methods} 
& Overall & 2D-count & 2D-relation & 3D-depth & 3D-distance
\\
\hline\hline

\rowcolor{gray!10} SpaceQwen2.5VL-3B 
& 51.4 & 62.2 & 45.4 & 45.4 & 50.0
\\
Kimi-VL-3B-Thinking
& 57.5 & 60.5 & 79.1 & 43.5 & 44.0
\\
\rowcolor{gray!10} Qwen2.5-VL-3B-Instruct
& 68.5 & 62.6 & 70.3 & 78.0 & 64.7
\\
InternVL2.5-4B
& 74.1 & 68.0 & 79.9 & 80.7 & 69.2
\\

\arrayrulecolor{gray!50}\hline
LLaVA-OneVision-7B
& 73.2 & 69.2 & 77.9 & 81.7 & 65.0
\\
\rowcolor{gray!10} Qwen2.5-VL-7B-Instruct
& 76.6 & 63.7 & 87.7 & 85.5 & 72.7
\\
Cambrain-8B
& 62.2 & 60.7 & 81.7 & 55.0 & 50.5
\\

\arrayrulecolor{gray!50}\hline
LLaMA-3.2V-11B
& 58.2 & 59.0 & 55.9 & 67.3 & 50.5
\\
\rowcolor{gray!10} LLaMA-3.2V-11B-CoT
& 72.8 & 59.1 & 78.9 & 78.8 & 78.0
\\
LLaVA-1.5-13B
& 58.2 & 56.1 & 57.2 & 66.8 & 53.3
\\
\rowcolor{gray!10} SpaceLLaVA-13B
& 58.2 & 56.1 & 57.2 & 66.8 & 53.3
\\

\arrayrulecolor{gray!50}\hline
Qwen2.5-VL-32B
& 79.7 & 68.9 & 80.8 & 86.5 & \textbf{85.8}
\\
\rowcolor{gray!10} LLaVA-OneVision-72B
& 79.7 & \textbf{70.2} & \textbf{89.2} & 82.5 & 79.0
\\

\hline
\rowcolor[HTML]{D7F6FF}
\multicolumn{6}{l}{\textcolor{blue!70}{\textit{Qwen2.5-VL-7B}}} 
\\
\rowcolor[HTML]{D7F6FF}
+ GASP - Correspondence ($\mathcal{L}_{corr}$)
& 79.4 & 68.0 & 88.1 & 86.6 & 78.6
\\
\rowcolor[HTML]{D7F6FF}
+ GASP - Full Model ($\mathcal{L}_{corr}$ + $\mathcal{L}_{depth}$)
& \textbf{79.8} & 68.2 & 88.3 & \textbf{87.3} & 79.2
\\

\hline
\end{tabular}
} 
\vspace{-10pt}
\captionsetup{font=small}

\end{table}

\subsection{Spatial Reasoning Benchmarks Evaluation}
\label{sec:benchmark_eval}

Our analysis (Section~\ref{sec:correspondence_analysis}) confirmed GASP improves \emph{internal} geometric representations. The ultimate goal is translating this to superior \emph{high-level} reasoning. We thus evaluate on downstream spatial VQA benchmarks to assess generalization for complex, language-based spatial tasks.

\noindent\textbf{Evaluation Benchmarks.}
Our primary evaluation uses benchmarks designed for 3D and multi-view spatial understanding: All-Angles Bench~\cite{yeh2025seeing} (varying viewpoints), VSI-Bench~\cite{yang2025thinking} (object permanence, relational reasoning), and the spatial subset of BLINK~\cite{fu2024blink} (relative depth, multi-view perception). We select these benchmarks as they are designed to isolate 'pure' geometric reasoning (e.g., viewpoint consistency) from the high-level semantic reasoning often confounded in 3D VQA datasets. The evaluation follows its standard protocol~\cite{yang2025thinking}. Second, to ensure our specialized training does not cause catastrophic forgetting, we evaluate on a suite of broad VQA benchmarks. This includes CV-Bench~\cite{tong2024cambrian} (2D/3D tasks), as well as Video-MME~\cite{fu2025video}, TempCompass~\cite{liu2024tempcompass}, and NextQA~\cite{xiao2021next} for holistic video and temporal understanding.

\noindent\textbf{Comparison Baselines.}
We compare our models against a wide range of state-of-the-art models including (1) \textit{General VLMs} (e.g., GPT-4o, InternVL2.5)~\cite{hurst2024gpto, team2023gemini, tong2024cambrian, chen2024internvl, bai2025qwen2, li2024llavaov, grattafiori2024llama, liu2024improved, team2025kimi} and (2) \textit{3D Spatial Reasoning VLMs} (e.g., VG-LLM, VLM-3R)~\cite{zheng2025learning, shi2025enhancing, fan2025vlm, chen2024spatialvlm}. To isolate the specific contribution of our geometric-aware training, our primary comparison is a controlled ablation. We establish our control models by fine-tuning the LLaVA-NeXT-Video-7B~\cite{zhang2024llavanextvideo} and Qwen2.5-VL-7B~\cite{bai2025qwen2} models on the general-purpose LLaVA-Video 178K dataset~\cite{zhang2024llavanextvideo}. Our models are trained on the exact same LLaVA-Video 178K dataset, but also include our proposed GASP geometric priors ($\mathcal{L}_{corr}$ and $\mathcal{L}_{depth}$). This setup ensures that all observed performance gains are directly attributable to our geometric training paradigm, not just to the effects of continued SFT. Furthermore, to disentangle the effect of data exposure from our training objective, we include a ``Fairness Baseline.'' Using the same DL3DV point tracks as GASP, we construct a VQA dataset where the correspondence task is reformulated as explicit question-answer pairs by following~\cite{zhang2025flatland, xu2025multi} (e.g., \textit{Which labeled point in Image-2 corresponds to the marked location in Image-1?} or \textit{Predict the $[x, y]$ coordinates of this point in Image-2.}). This baseline is fine-tuned on the identical data mix (LLaVA-Video 178K + DL3DV VQA) with the same data quantity, isolating whether gains arise from data content or from GASP's geometric objective.

\subsection{Benchmark Results}
\label{sec:spatial_results}

Our GASP framework improves over the baselines on several spatial benchmarks in Table~\ref{tab:main_spatial_results}, with the largest gains on tasks directly related to the injected geometric priors. This provides evidence that learning from fundamental 3D priors can enhance high-level spatial reasoning. We highlight our findings as follows:

\begin{table}[t]
\centering
\captionsetup{font=small}
\caption{\textbf{Comparison on generic multimodal benchmarks.}}
\label{tab:multimodal_benchmarks}
\vspace{-10pt}
\resizebox{\columnwidth}{!}{
\begin{tabular}{lcccc}
\hline \hline
\rowcolor{gray!20}
Methods & 
\rotatebox{30}{\begin{tabular}[c]{@{}c@{}}\textbf{Video-MME}\\ \textbf{w/o sub.}\end{tabular}} & 
\rotatebox{30}{\begin{tabular}[c]{@{}c@{}}\textbf{Video-MME}\\ \textbf{w/ sub.}\end{tabular}} & 
\rotatebox{30}{\begin{tabular}[c]{@{}c@{}}\textbf{Temp}\\ \textbf{Compass\textsubscript{MC}}\end{tabular}}  & 
\rotatebox{30}{\textbf{NextQA}} \\
\midrule
\rowcolor[HTML]{D7F6FF}
\multicolumn{5}{l}{\textcolor{blue!70}{\textit{LLaVA-NeXT-Video-7B}}} \\
\rowcolor[HTML]{D7F6FF}
(baseline) & 40.8 & 40.3 & 50.1 & 59.4 \\
\rowcolor[HTML]{D7F6FF}
+ GASP - Correspondence  & 42.3 & 41.6 & 53.7 & 62.8 \\
\rowcolor[HTML]{D7F6FF}
+ GASP - Full Model  & 42.8 & 41.9 & 53.8 & 61.6 \\
\midrule 
\rowcolor[HTML]{D7F6FF}
\multicolumn{5}{l}{\textcolor{blue!70}{\textit{Qwen2.5-VL-7B}}} \\
\rowcolor[HTML]{D7F6FF}
(baseline) & 60.6 & 59.3 & 68.4 & 76.6 \\
\rowcolor[HTML]{D7F6FF}
+ GASP - Correspondence  & 62.6 & 61.2 & 71.5 & 78.4 \\
\rowcolor[HTML]{D7F6FF}
+ GASP - Full Model  & 63.2 & 61.6 & 70.3 & 74.7 \\
\bottomrule
\end{tabular}
} 
\vspace{-10pt}

\end{table}

\noindent\textbf{Analysis of 3D Geometric Consistency.}
As shown in Table~\ref{tab:main_spatial_results}, our gains are most pronounced on the specific sub-tasks that are directly related to the geometric priors we inject. On the All-Angles Bench, which tests geometric consistency, our full model nearly doubles the Camera Pose Estimation score (e.g., 34.1\% to 52.8\% for Qwen2.5-VL) and improves Relative Direction (e.g., 24.7\% to 29.8\% for LLaVA-NeXT). This directly validates that our training imbues the model with a robust understanding of viewpoint. Also, the DL3DV VQA baseline even degrades several key metrics compared to the SFT baseline (e.g., Camera Pose 22.7\%$\rightarrow$19.8\%, Object Counting 23.5\%$\rightarrow$21.4\% for LLaVA-NeXT). In contrast, GASP achieves 40.9\% and 52.5\% on these same metrics. This confirms that the improvements stem from the GASP geometric objective rather than data exposure, consistent with the overfitting patterns of VQA-based methods observed in Appendix.

\noindent\textbf{Analysis of Relational \& Temporal Capability.}
This geometric understanding also improves abstract reasoning. On VSI-Bench, our model shows a dramatic improvement in Object Counting (e.g., 23.5\% to 52.5\% on VSI-Bench). This suggests that the view-invariant features learned by our method help the model maintain object identity, preventing it from double-counting or missing objects across frames. This multi-view consistency is further validated on BLINK, where our framework provides the boost on the Multi-View task (e.g., 42.1\% to 57.1\% for LLaVA-NeXT), proving our model's superior ability to correlate information across different perspectives.

\noindent\textbf{Analysis on General-Purpose Benchmarks.}
A natural concern is whether specialized geometric training harms general VQA capabilities. As presented in Table~\ref{tab:multimodal_benchmarks}, we observe a modest trade-off: our Qwen2.5-VL-7B model with GASP loses 1.9\% on NextQA (76.6\% $\rightarrow$ 74.7\%), but improves on temporal video benchmarks, including Video-MME (59.3\% $\rightarrow$ 61.6\%) and TempCompass (68.4\% $\rightarrow$ 70.3\%). We attribute these temporal gains to improved object permanence: video understanding requires maintaining persistent object identity across viewpoint changes, occlusions, and temporal gaps (e.g., \textit{Temporal Reasoning} and \textit{Action Forecasting} in Video-MME), which directly benefits from our view-invariant geometric representations learned through correspondence training.
The modest drop on NextQA (1.9\%) likely reflects a capacity trade-off: geometric specialization comes at a small cost to action-understanding tasks, which rely more on object semantics and temporal dynamics than on precise spatial localization. This suggests that GASP is best suited for applications where spatial geometry is primary (e.g., robotics, 3D reasoning) rather than action-centric understanding. Also, on the broad CV-Bench (Table~\ref{tab:cvbench_comparison}), our model achieves an Overall score of 79.8\%. Overall, GASP trades about 1--2\% general VQA accuracy for consistent gains on spatial and temporal benchmarks.

\begin{table}[t]
\centering
\caption{\textbf{Ablation studies of the LoRA rank effect and correspondence head injection into LLM layers.} }
\label{tab:lora_rank_ablation}
\vspace{-10pt}
\scriptsize
\resizebox{\columnwidth}{!}{
\begin{tabular}{lcccc}
\toprule

\makecell[l]{\textbf{GASP - Full Model} \\ \textbf{($\mathcal{L}_{corr}$ + $\mathcal{L}_{depth}$)}} 
& \textbf{Avg. PCK} & \textbf{All-Angles Bench} & \textbf{VSI-Bench} & \textbf{BLINK} \\
\midrule
\multicolumn{5}{c}{\textcolor{gray!110}{\textit{Impact of LoRA Rank on Performance}}} \\

\rowcolor[HTML]{D7F6FF} \multicolumn{5}{l}{\textcolor{blue!70}{\textit{LLaVA-NeXT-Video-7B}}} \\
\rowcolor[HTML]{D7F6FF} LoRA Rank = 64   & 8.4  & 30.1 & 28.5 & 44.9 \\
\rowcolor[HTML]{D7F6FF} LoRA Rank = 128  & 13.7 & 32.6 & 30.6 & 45.8 \\
\rowcolor[HTML]{D7F6FF} LoRA Rank = 256  & 17.1 & 35.8 & 33.9 & 47.5 \\
\rowcolor[HTML]{D7F6FF} LoRA Rank = 512  & 26.2 & \textbf{38.1} & \textbf{37.1} & \textbf{51.0} \\
\rowcolor[HTML]{D7F6FF} LoRA Rank = 1024 & \textbf{28.6} & 37.2 & 34.8 & 48.7 \\
\midrule
\rowcolor[HTML]{D7F6FF} \multicolumn{5}{l}{\textcolor{blue!70}{\textit{Qwen2.5-VL-7B}}} \\
\rowcolor[HTML]{D7F6FF} LoRA Rank = 64   & 18.2 & 38.5 & \textbf{37.3} & 70.2 \\
\rowcolor[HTML]{D7F6FF} LoRA Rank = 128  & 26.7 & \textbf{43.4} & 36.9 & \textbf{74.3} \\
\rowcolor[HTML]{D7F6FF} LoRA Rank = 256  & 28.8 & 41.8 & 35.5 & 73.5 \\
\rowcolor[HTML]{D7F6FF} LoRA Rank = 512  & 31.2 & 40.2 & 34.1 & 72.4 \\
\rowcolor[HTML]{D7F6FF} LoRA Rank = 1024 & \textbf{32.5} & 38.9 & 33.2 & 73.8 \\
\hline
\midrule

\multicolumn{5}{c}{\textcolor{gray!110}{\textit{Correspondence Head Injection into LLM Layers}}} \\
\rowcolor[HTML]{D7F6FF} \multicolumn{5}{l}{\textcolor{blue!70}{\textit{LLaVA-NeXT-Video-7B}}} \\

\rowcolor[HTML]{D7F6FF} Layer 10 - 18      & 21.7 & 34.8 & 35.9 & 47.7 \\
\rowcolor[HTML]{D7F6FF} Layer 18 - 25      & 25.1 & 37.5 & 35.2 & 49.5 \\
\rowcolor[HTML]{D7F6FF} Layer 25 - 32      & 25.8 & \textbf{39.1} & 36.5 & 49.3 \\
\rowcolor[HTML]{D7F6FF} All Layers (1 - 32) & \textbf{26.2} & 38.1 & \textbf{37.1} & \textbf{51.0} \\

\midrule
\rowcolor[HTML]{D7F6FF} \multicolumn{5}{l}{\textcolor{blue!70}{\textit{Qwen2.5-VL-7B}}} \\
\rowcolor[HTML]{D7F6FF} Layer 10 - 16      & 19.8 & 37.9 & 34.2 & 68.2 \\
\rowcolor[HTML]{D7F6FF} Layer 16 - 22      & 23.3 & 38.8 & 35.5 & 71.1 \\
\rowcolor[HTML]{D7F6FF} Layer 22 - 28      & 25.2 & 42.7 & \textbf{37.4} & 72.8 \\
\rowcolor[HTML]{D7F6FF} All Layers (1 - 28) & \textbf{26.7} & \textbf{43.4} & 36.9 & \textbf{74.3} \\

\bottomrule
\end{tabular}
}
\vspace{-10pt}
\end{table}

\subsection{Ablation Studies}
\label{sec:ablation_studies}

We conduct ablation studies on two key hyperparameters for our GASP framework: the LoRA rank and the correspondence head ($\mathcal{H}_c$) injection strategy, reported in Table~\ref{tab:lora_rank_ablation}.

\noindent\textbf{Impact of LoRA Rank.}
We analyze performance varying the LoRA rank from 64 to 1024. We find a clear trade-off: internal correspondence (Avg. PCK) generally scales with the rank, but downstream performance peaks earlier (512 for LLaVA, 128 for Qwen). This suggests that a higher Avg. PCK does not always lead to improved complex spatial benchmark reasoning. We hypothesize that with very high LoRA ranks, while it fits better for our geometric priors, may begin to harm foundational language capabilities.

\noindent\textbf{Correspondence Head Injection into LLM Layers.}
We target LLM layers rather than the visual encoder because spatial reasoning fundamentally requires sequence-level temporal binding. We then ablate \emph{where} in the LLM the correspondence head is injected. While injecting at later layers (e.g., LLaVA `Layer 25-32`), which contain richer semantic information, outperforms earlier layers, our key finding is that applying supervision across all layers (1-32 for LLaVA, 1-28 for Qwen) yields the best and most consistent performance. This result suggests that geometric consistency is fundamentally hierarchical: early layers must learn to match low-level visual features (edges, corners), middle layers must reason about object parts and boundaries, and deep layers must maintain semantic-geometric alignment. By supervising at all levels, we ensure that gradients from the geometric losses flow throughout the network, forcing each layer to contribute to view-invariant feature learning. If geometric supervision were only applied at deep layers, shallow layers might continue to learn view-dependent features, creating a representational bottleneck.

\section{Conclusion}
\label{sec:conclusion}

In this paper, we proposed GASP, a framework that injects fundamental geometric priors directly into the LLM's transformer layers. Our analysis showed GASP corrects VLMs' near-zero internal correspondence accuracy, boosting it to over 70\%. These internal improvements generalize, achieving significant gains on downstream spatial benchmarks. We thus find learning from geometric priors is a promising and generalizable path toward spatially-intelligent VLMs. Current limitations include reliance on pseudo ground-truth depth and a modest trade-off on action-centric tasks; future work could combine geometric priors with complementary VQA supervision and scale to larger model architectures.

{
    \small
    \bibliographystyle{ieeenat_fullname}
    \bibliography{main}
}

\clearpage
\beginappendix

\section*{Overview}
In this supplementary material, we provide details on our geometric training data collection in Section~\ref{sec:supp-data-viz}. Next, we provide full implementation details, including the correspondence head architecture ($\mathcal{H}_{c}$) and all training hyperparameters, in Section~\ref{sec:supp-implementation}. Following this, we detail the evaluation protocol used to measure correspondence in both our model and the baselines in Section~\ref{sec:supp-baseline-protocol}. We then provide a quantitative analysis of the VSI-Bench dataset, exploring its inherent biases and the performance of SFT-trained models in Section~\ref{sec:supp-vsi-bias}. We subsequently provide a brief theoretical overview of the gradient backpropagation from our geometric losses in Section~\ref{sec:supp-gradient-flow}. Finally, we discuss the fundamental distinction between our learned geometric correspondence and standard rotary positional embeddings (RoPEs) in Section~\ref{sec:supp-pe-discussion}.

\section{Training Dataset Collection}
\label{sec:supp-data-viz}

We leverage the multi-view video sequences and depth maps from DL3DV~\cite{ling2024dl3dv} and follow the VGGT's annotation recipe~\cite{wang2025vggt} to generate dense point correspondence annotations for training. 

For each scene, we use the provided camera intrinsics $K \in \mathbb{R}^{3 \times 3}$ and extrinsics $[R|t] \in \mathbb{R}^{3 \times 4}$ from COLMAP's Structure-from-Motion reconstruction in DL3DV~\cite{ling2024dl3dv} and VGGT~\cite{wang2025vggt}. Given a query frame (frame 0) with depth map $D_0 \in \mathbb{R}^{H \times W}$, we back-project valid pixels to 3D world coordinates using $\mathbf{p}_w = K^{-1} D_0(u,v) [u, v, 1]^T$, where $(u,v)$ denotes the pixel coordinate. These world points are then projected to subsequent frames using $\mathbf{p}_i = K [R_i | t_i] \mathbf{p}_w$ to establish correspondences. We validate each correspondence through depth consistency: a projected point is considered valid only if the depth difference satisfies $|D_{proj} - D_{map}| < 0.05 \times \min(D_{proj}, D_{map})$, where $D_{proj}$ is the projected depth and $D_{map}$ is the depth map value at the projected location. Also, we enforce a boundary margin of 4 pixels from image edges to avoid projection artifacts.

To construct a balanced training signal, we sample both positive and negative correspondences. Positive tracks are sampled from validated 3D projections, prioritizing points that remain visible across multiple frames (at least 2 frames). We target $8\times8$ and $24\times24$ points per video frame and retain the top 50\% of tracks ranked by visibility duration. Negative samples are generated by applying random spatial perturbations (within 50\%).

\section{Additional Implementation Details}
\label{sec:supp-implementation}

Here, we provide the specific architectural and training details required for reproducibility.

\noindent\textbf{Correspondence Head ($\mathcal{H}_{c}$) Architecture.}
The correspondence head $\mathcal{H}_{c}$ is implemented as a 2-layer MLP consisting of a Linear layer that projects from hidden dimension $d_h$ to $d_h/2$, followed by GELU activation, and a second Linear layer projecting back to $d_h$. For our experiments, $d_h = 3584$ for Qwen2.5-VL-7B~\cite{bai2025qwen2} and $d_h = 4096$ for LLaVA-NeXT-Video-7B~\cite{zhang2024llavanextvideo}. The head is initialized using SVD decomposition of the query projection matrix ($\mathbf{W}_Q$) from the corresponding attention layer.

\noindent\textbf{Training Hyperparameters.}
We employ LoRA fine-tuning with rank $r=512$ for LLaVA-NeXT-Video-7B and $r=128$ for Qwen2.5-VL-7B, applied only to attention projection matrices ($W_Q, W_K, W_V, W_O$). The correspondence head is trained with full precision. We use cosine learning rate scheduling with 10\% warmup over 3 epochs. For the loss function (Equation~\ref{eq:total_loss}), we set the contrastive loss weight $\lambda_c = 0.3$ and distance loss weight $\lambda_d = 1.0$.

\noindent\textbf{Joint Training Data Composition.}
Our joint training combines the DL3DV-derived 3D scene dataset (1.75M point correspondence annotations) with LLaVA-Video-178K (100K general video QA samples). This composition ensures the model maintains strong general video understanding capabilities while acquiring fine-grained spatial reasoning abilities.

\begin{table*}[t]\small
\centering
\caption{\textbf{Analysis of VSI-Bench dataset bias.} We compare the baseline models against themselves when provided with the dataset's average object and room sizes as a textual "prior" in the prompt. Deltas for VLM-3R are shown relative to the LLaVA-NeXT-Video Baseline (7B\&72B).}
\label{tab:bias-hack}
\vspace{-10pt}
\scriptsize{
\resizebox{\linewidth}{!}{ 
\renewcommand\arraystretch{1.1} 
\setlength\tabcolsep{3.pt}
\begin{tabular}{llccccc} 
\hline\hline
\rowcolor{gray!20}
\textbf{Task} & \textbf{Metric} & \textbf{Baseline (7B)} & \textbf{Baseline (7B) + Avg. Prior} & \textbf{Baseline (72B)} & \textbf{Baseline (72B) + Avg. Prior} & \textbf{VLM-3R} \\
\hline\hline
Object Size Estimation & MRA@.5:.95:.05 & 0.47 & 0.64 (\textbf{$\Delta$ +0.17}) & 0.57 & 0.65 (\textbf{$\Delta$ +0.08}) & 0.69 (\textbf{$\Delta$ +0.22}) \\
Room Size Estimation & MRA@.5:.95:.05 & 0.24 & 0.38 (\textbf{$\Delta$ +0.14}) & 0.36 & 0.46 (\textbf{$\Delta$ +0.10}) & 0.67 (\textbf{$\Delta$ +0.43}) \\
Object Abs Distance & MRA@.5:.95:.05 & 0.14 & 0.61 (\textbf{$\Delta$ +0.47}) & 0.23 & 0.57 (\textbf{$\Delta$ +0.34}) & 0.49 (\textbf{$\Delta$ +0.36}) \\
\hline\hline
\end{tabular}
} %
} %
\end{table*}

\section{Correspondence Evaluation Protocol}
\label{sec:supp-baseline-protocol}

This section details the exact methodology used to compute correspondence accuracy (PCK) for both baseline models (LLaVA-NeXT-Video-7B, Qwen2.5-VL-7B) and our GASP models. For baseline models lacking explicit correspondence heads, we extract query states $Q$ and key states $K$ from each transformer layer during forward pass. Visual tokens are isolated by slicing the sequence from position $T_s$ to $T_e$ where $T_s$ denotes the first visual token position and $T_e = T_s + N_f \times N_p$ with $N_f$ being the number of frames and $N_p$ the patches per frame. The extracted features are reshaped to $[N_f, N_p, d_h]$ where $d_h$ is the hidden dimension. For models employing Grouped-Query Attention (GQA), we average over attention heads to obtain $[N_p, \bar{d}]$ where $\bar{d} = d_h / n_h$. Given source frame features $Q_0 \in \mathbb{R}^{N_p \times \bar{d}}$ and target frame features $K_j \in \mathbb{R}^{N_p \times \bar{d}}$ for frame $j$, we compute the correspondence matrix using cosine similarity: $S = \text{CosineSim}(Q_0, K_j^T)$, and the predicted target patch for source patch $i$ is $\hat{p}_i = \arg\max_j S_{ij}$.

We convert both ground-truth and predicted patch indices to 2D grid coordinates and compute the Euclidean distance $d = \|(r_{gt}, c_{gt}) - (r_{pred}, c_{pred})\|_2$ in patch space. We separately compute confidence on correct predictions ($d < 2$) versus incorrect predictions to obtain the calibration gap, which measures whether the model exhibits awareness of its prediction quality.

\section{Analysis of VSI-Bench Dataset Bias}
\label{sec:supp-vsi-bias}

A potential criticism of high performance on benchmarks like VSI-Bench is that models may "hack" the benchmark by learning superficial dataset-specific biases (e.g., "all microwaves are 0.5m wide") rather than performing genuine 3D reasoning.

\noindent\textbf{Bias Hacking Experiment.}
To investigate the extent to which VSI-Bench scores can be "hacked" by exploiting dataset-level biases, we conducted an experiment using a text-based prior. We first quantified these biases by extracting the object and room sizes from the VSI-bench QAs and averaging them. This yielded a dictionary of average object sizes (e.g., \texttt{'sofa': 181.30}, \texttt{'bed': 216.06}) and an average room size of 20.5 square meters.

Instead of a "bias-only" model, we provided these averaged values directly to the baseline VLMs as part of the input prompt, e.g., \textit{``The average room size is 20.5 square meters. Use this information to guide your estimate.''} As shown in Table~\ref{tab:bias-hack}, this simple textual prior dramatically boosts performance. For example, the LLaVA-NeXT-Video-7B baseline's "Object Abs Distance" score skyrockets from 0.14 to 0.61 (+0.47), and the LLaVA-NeXT-Video-72B model's score jumps from 0.23 to 0.57 (+0.34). Notably, on this task, the baseline models with this simple prior (0.61 and 0.57) both significantly outperform the SFT-trained VLM-3R (0.49). This finding indicates that \textit{a significant portion of the benchmark's challenge can be solved by exploiting these easily-averaged dataset statistics, rather than relying solely on complex, visual-based spatial reasoning.}

Our observation mirrors the recent findings~\cite{brown2025benchmark} where they demonstrated that VSI-Bench contains pervasive non-visual shortcuts that allow models to bypass genuine visual reasoning. Their diagnostic ``Test-set Stress-Test'' revealed that statistical regularities in the answer distribution enable high performance even without visual input, a vulnerability our experiment empirically validates by explicitly exploiting these statistical priors.

\begin{figure}[!t]
    \centering
    \includegraphics[width=\linewidth]{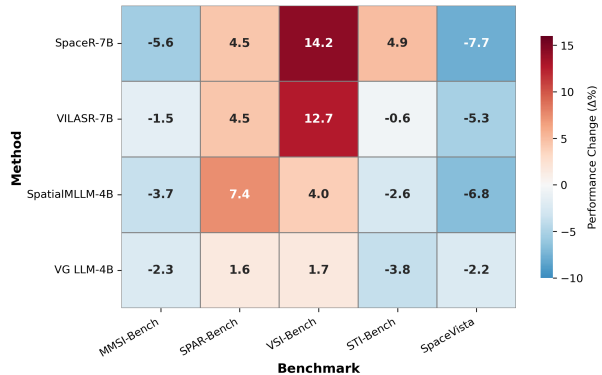}
    \caption{\textbf{Generalization Gap in 3D-VQA Fine-Tuning.} We illustrate the performance change ($\Delta\%$) of specialized spatial VLMs relative to their underlying pre-trained backbones across five distinct spatial benchmarks. While fine-tuning yields significant improvements on specific datasets (e.g., VSI-Bench, highlighted in red), it consistently leads to performance degradation (blue cells) on out-of-distribution benchmarks like MMSI-Bench and SpaceVista. This performance profile suggests that standard SFT strategies suffer from severe overfitting to dataset-specific biases, whereas genuine spatial understanding should generalize across domains.}
    \label{fig:supp-generalization}
\end{figure}

\noindent\textbf{Generalization Analysis of 3D-VQA Models.}
To empirically validate the generalization limitations of standard 3D-VQA fine-tuning, we conduct a cross-dataset performance analysis in Figure~\ref{fig:supp-generalization}. We report the relative performance change ($\Delta\%$) of state-of-the-art spatial VLMs compared to their respective pre-trained base models (e.g., Qwen2.5-VL). A clear pattern of \textit{task-specific overfitting} emerges: models like SpaceR-7B~\cite{ouyang2025spacer} and VILASR-7B~\cite{wu2025reinforcing} achieve substantial gains on VSI-Bench~\cite{yang2025thinking} (+14.2\% and +12.7\%), likely due to high similarity between their training mixtures and this specific benchmark. 

However, this comes at the cost of negative transfer on other spatial benchmarks. Notably, performance degrades significantly on MMSI-Bench~\cite{yang2025mmsi}, STI-Bench~\cite{li2025sti}, and SpaceVista~\cite{sun2025spacevista} (dropping by as much as -7.7\%), indicating that these models are memorizing dataset-specific distributions rather than acquiring robust, generalized spatial reasoning. This stark contrast underscores the necessity of our GASP approach, which injects fundamental geometric priors to avoid such brittle memorization.

\section{Analysis of Gradient Backpropagation}

\label{sec:supp-gradient-flow}

The total loss for our framework is $\mathcal{L}_{total} = \mathcal{L}_{LM} + \lambda_{c}\mathcal{L}_{corr} + \lambda_{d}\mathcal{L}_{depth}$. The key to our method is how the geometric-aware gradients from $\mathcal{L}_{corr}$ and $\mathcal{L}_{depth}$ backpropagate through the correspondence head to update the backbone's parameters, specifically the Query ($Q$) and Key ($K$) projectors within the transformer layers.

Formally, let $\theta^{(l)} = \{W_Q^{(l)}, W_K^{(l)}, W_V^{(l)}\}$ denote the learnable weights of the Self-Attention mechanism at transformer layer $l$. The visual tokens $V^{(l)}$ output by this layer serve as the input to our lightweight correspondence head $\mathcal{H}_c$. The gradient of the total loss with respect to the backbone weights $\theta^{(l)}$ can be decomposed as:

\begin{equation}
    \frac{\partial \mathcal{L}_{total}}{\partial \theta^{(l)}} = \underbrace{\frac{\partial \mathcal{L}_{LM}}{\partial \theta^{(l)}}}_{\text{Language Modeling}} + \underbrace{\lambda_c \frac{\partial \mathcal{L}_{corr}}{\partial \theta^{(l)}} + \lambda_d \frac{\partial \mathcal{L}_{depth}}{\partial \theta^{(l)}}}_{\text{Geometric Supervision}}
\end{equation}

\noindent We focus on the geometric term. Since the correspondence embeddings are defined as $E = \mathcal{H}_c(V^{(l)})$ (Equation~\ref{eq:correspondence_head}), the gradients flow via the chain rule:

\begin{equation}
    \frac{\partial \mathcal{L}_{corr}}{\partial \theta^{(l)}} = \frac{\partial \mathcal{L}_{corr}}{\partial E} \cdot \frac{\partial \mathcal{H}_c(V^{(l)})}{\partial V^{(l)}} \cdot \frac{\partial V^{(l)}}{\partial \theta^{(l)}}
\end{equation}

The term $\frac{\partial V^{(l)}}{\partial \theta^{(l)}}$ acts as a \textit{Gradient Bridge}. Recall that self-attention is defined as $Z = \text{Softmax}(\frac{QK^T}{\sqrt{d_k}})V = A \cdot V$, where $Q = X^{(l-1)}W_Q^{(l)}$, $K = X^{(l-1)}W_K^{(l)}$, $V = X^{(l-1)}W_V^{(l)}$, and $A = \text{Softmax}(\frac{QK^T}{\sqrt{d_k}})$. The output visual tokens are $V^{(l)} = Z + X^{(l-1)}$. Applying the chain rule through the attention mechanism:

\begin{equation}
    \frac{\partial V^{(l)}}{\partial W_Q^{(l)}} = \frac{\partial (A \cdot V)}{\partial A} \cdot \frac{\partial A}{\partial (QK^T)} \cdot \frac{\partial (QK^T)}{\partial Q} \cdot \frac{\partial Q}{\partial W_Q^{(l)}}
\end{equation}

\noindent The key components are: $\frac{\partial Q}{\partial W_Q^{(l)}} = (X^{(l-1)})^T$, $\frac{\partial (QK^T)}{\partial Q} = K$, the softmax Jacobian $\frac{\partial A_{ij}}{\partial S_{kl}} = A_{ij}(\delta_{ik}\delta_{jl} - A_{il})$ where $S = \frac{QK^T}{\sqrt{d_k}}$, and $\frac{\partial (A \cdot V)}{\partial A} = V^T$. Combining these, the gradient with respect to $W_Q^{(l)}$ becomes:

\begin{equation}
    \frac{\partial \mathcal{L}_{corr}}{\partial W_Q^{(l)}} = (X^{(l-1)})^T \cdot \left[\frac{1}{\sqrt{d_k}} K \cdot \nabla_A^{\text{softmax}} \cdot V \cdot \frac{\partial \mathcal{L}_{corr}}{\partial V^{(l)}}\right]
\end{equation}

\noindent where $\nabla_A^{\text{softmax}} = \text{diag}(A)(I - \mathbf{1}A)$ is the softmax gradient term. Similarly, for $W_K^{(l)}$:

\begin{equation}
    \frac{\partial \mathcal{L}_{corr}}{\partial W_K^{(l)}} = (X^{(l-1)})^T \cdot \left[\frac{1}{\sqrt{d_k}} Q^T \cdot \nabla_A^{\text{softmax}} \cdot V \cdot \frac{\partial \mathcal{L}_{corr}}{\partial V^{(l)}}\right]
\end{equation}

\noindent\textbf{Geometric Gradient Structure.} The correspondence loss $\mathcal{L}_{corr}$ is a contrastive objective over correspondence embeddings. For frames $(I_t, I_{t'})$ with matched points $(p_i, p_j)$ and embeddings $(e_i, e_j)$:

\begin{equation}
    \mathcal{L}_{corr} = -\log \frac{\exp(\text{sim}(e_i, e_j)/\tau)}{\sum_{k \in \mathcal{N}} \exp(\text{sim}(e_i, e_k)/\tau)}
\end{equation}

\noindent where $\mathcal{N}$ includes positive and negative samples. The derivative is:

\begin{equation}
    \frac{\partial \mathcal{L}_{corr}}{\partial e_i} = \frac{1}{\tau}\left[\sum_{k \in \mathcal{N}} p_k \cdot \frac{\partial \text{sim}(e_i, e_k)}{\partial e_i} - \frac{\partial \text{sim}(e_i, e_j)}{\partial e_i}\right]
\end{equation}

\noindent where $p_k = \frac{\exp(\text{sim}(e_i, e_k)/\tau)}{\sum_{l} \exp(\text{sim}(e_i, e_l)/\tau)}$. This gradient pushes $e_i$ towards its correspondence $e_j$ while pulling away from negatives, creating view-invariance. Crucially, backpropagating through $\mathcal{H}_c$:

\begin{equation}
    \frac{\partial \mathcal{L}_{corr}}{\partial V^{(l)}} = \left(\frac{\partial \mathcal{H}_c}{\partial V^{(l)}}\right)^T \cdot \frac{\partial \mathcal{L}_{corr}}{\partial E}
\end{equation}

\noindent produces a spatially localized gradient that differs fundamentally from the dense semantic gradient $\frac{\partial \mathcal{L}_{LM}}{\partial V^{(l)}}$. This teaches the attention mechanism to distinguish tokens by 3D spatial positions, not just semantic categories.

\noindent\textbf{Impact on Query-Key Similarity.} The similarity between tokens $i$ and $j$ is:

\begin{equation}
    S_{ij} = \frac{q_i^T k_j}{\sqrt{d_k}} = \frac{x_i^T W_Q^T W_K x_j}{\sqrt{d_k}}
\end{equation}

\noindent The gradient update due to $\mathcal{L}_{corr}$ is:

\begin{equation}
    \Delta S_{ij} = -\eta \lambda_c \frac{\partial \mathcal{L}_{corr}}{\partial S_{ij}} = -\eta \lambda_c \left[\frac{\partial \mathcal{L}_{corr}}{\partial V^{(l)}} \cdot \frac{\partial V^{(l)}}{\partial A} \cdot \frac{\partial A}{\partial S_{ij}}\right]
\end{equation}

\noindent where $\eta$ is the learning rate. This update increases $S_{ij}$ for spatially corresponding tokens and decreases it for geometrically distinct tokens, even if semantically similar. Over training, the projector product $W_Q^T W_K$ learns to encode geometric correspondence:

\begin{equation}
    W_Q^{T,(l)} W_K^{(l)} \approx M_{\text{geo}} + M_{\text{sem}}
\end{equation}

\noindent where $M_{\text{geo}}$ encodes geometric alignment (high values for corresponding 3D locations) and $M_{\text{sem}}$ encodes semantic similarity (from $\mathcal{L}_{LM}$). The geometric term emerges from the accumulated gradients:

\begin{equation}
    M_{\text{geo}} = \sum_{t=1}^{T} \eta \lambda_c \left[\frac{\partial \mathcal{L}_{corr}}{\partial W_Q^{(l)}}\right]^T \left[\frac{\partial \mathcal{L}_{corr}}{\partial W_K^{(l)}}\right]
\end{equation}

\noindent\textbf{Depth Consistency Regularization.} The depth loss $\mathcal{L}_{depth} = \sum_{i,j} A_{ij} \cdot \mathcal{D}(d_i, d_j)$ penalizes depth-inconsistent matches, where $\mathcal{D}(\cdot, \cdot)$ measures depth discrepancy. The gradient is:

\begin{equation}
    \frac{\partial \mathcal{L}_{depth}}{\partial A_{ij}} = \mathcal{D}(d_i, d_j)
\end{equation}

\noindent Backpropagating through softmax:

\begin{equation}
    \frac{\partial \mathcal{L}_{depth}}{\partial S_{ij}} = \mathcal{D}(d_i, d_j) \cdot A_{ij}(1 - A_{ij})
    \label{eq:depth_grad_softmax}
\end{equation}

\noindent The term $A_{ij}(1 - A_{ij})$ amplifies gradients for mid-confidence predictions ($A_{ij} \approx 0.5$), teaching the model to suppress geometrically invalid matches. This creates depth-aware projectors:

\begin{equation}
    \frac{\partial \mathcal{L}_{depth}}{\partial W_Q^{(l)}} = (X^{(l-1)})^T \cdot \left[\frac{1}{\sqrt{d_k}} K \cdot \text{diag}(\mathcal{D}) \cdot \nabla_A^{\text{softmax}} \cdot V\right]
\end{equation}

\noindent where $\text{diag}(\mathcal{D})$ is a diagonal matrix of depth discrepancies. This modulates the attention mechanism to respect 3D boundaries, effectively learning:

\begin{equation}
    S_{ij}^{\text{effective}} = \frac{x_i^T W_Q^T W_K x_j}{\sqrt{d_k}} - \lambda_d \cdot \mathcal{D}(d_i, d_j) + \text{noise}
\end{equation}

\noindent where the depth penalty is implicitly encoded in $W_Q^T W_K$.

\noindent\textbf{QK Enhancement Mechanism.} The correspondence head creates two synergistic effects. First, \textit{Geometric Subspace Alignment}: the gradient update

\begin{equation}
    W_Q^{(l)} \leftarrow W_Q^{(l)} - \eta \lambda_c \frac{\partial \mathcal{L}_{corr}}{\partial W_Q^{(l)}}
\end{equation}

\noindent incorporates $K \cdot \nabla_A^{\text{softmax}} \cdot V \cdot \frac{\partial \mathcal{L}_{corr}}{\partial V^{(l)}}$ (from Equation~\ref{eq:correspondence_head}), which couples the current Key representations with geometric error signals. Over iterations, $W_Q$ and $W_K$ co-evolve:

\begin{equation}
    \langle W_Q^{(l)} x_i, W_K^{(l)} x_j \rangle \to \max \quad \text{if } (x_i, x_j) \text{ corresponds}
\end{equation}

\noindent Second, \textit{Depth-Aware Pruning}: the depth gradient (Equation~\ref{eq:depth_grad_softmax}) forces attention weights to respect 3D structure. The combined effect yields learned attention weights:

\begin{equation}
    A_{ij}^{\text{learned}} = \text{Softmax}\left(\frac{x_i^T W_Q^T W_K x_j}{\sqrt{d_k}}\right)
\end{equation}

\noindent that are high for geometrically corresponding and depth-consistent token pairs, and low otherwise. Consequently, although $\mathcal{H}_c$ is discarded at inference, these geometric priors are permanently baked into $\theta^{(l)}$. The learned projectors $W_Q^{(l)}$ and $W_K^{(l)}$ encode: (1) spatial correspondence, where tokens at corresponding 3D locations produce high $S_{ij}$; (2) view invariance, where the QK space is invariant to perspective/lighting changes; (3) depth awareness, where attention respects 3D scene structure. This enables the standard VLM to perform robust spatial reasoning without auxiliary inputs, as the attention mechanism itself has been geometrically restructured. The correspondence head guides the backbone to internalize 3D-aware attention patterns.

\section{Relation to Positional Embeddings}
\label{sec:supp-pe-discussion}

\noindent\textbf{Rotary Positional Embeddings.}
Standard Vision Transformers (ViTs) and VLMs utilize Positional Embeddings (PEs), such as absolute learnable embeddings~\cite{dosovitskiy2020image} or Rotary Positional Embeddings (RoPE)~\cite{su2024roformer}, to inject grid location information into the sequence. Similarly, Video Transformers often extend this to 3D-RoPEs~\cite{arnab2021vivit, bertasius2021space} by adding a temporal or depth dimension.
However, these RoPEs provide only \textit{static coordinate information} (e.g., "this token is at location $(x,y)$"). They do not encode \textit{visual correspondence} or \textit{object permanence}. As evidenced in our main paper (Figure~\ref{fig:analysis_grid}), the baseline models (Qwen2.5-VL and LLaVA-NeXT), which are already equipped with advanced RoPE, achieve near-zero correspondence accuracy. This empirically demonstrates that providing coordinate information via RoPE is insufficient for the model to learn that an object at location $(x_1, y_1)$ in Frame $t$ is the same entity as the one at $(x_2, y_2)$ in Frame $t+1$.

\noindent\textbf{Our GASP: From Coordinates to Correspondence.}
In contrast to RoPE, which is an \textit{input-level} signal, GASP operates on the \textit{interaction mechanism} ($QK^T$) of the model.
\begin{itemize}
    \item \textbf{Content-Aware vs. Location-Aware:} RoPE is content-agnostic; it is identical for a blank wall or a complex face. GASP, supervised by our contrastive loss $\mathcal{L}_{corr}$, forces the visual features to be \textit{content-aware}. It ensures that the query representation of an object matches its key representation in another view, regardless of their disparate positional encodings.
    \item \textbf{Implicit 3D Consistency vs. Explicit 3D Input:} While approaches like 3D-RoPE require explicit 3D inputs (e.g., depth maps or point clouds) to encode geometry, GASP internalizes 3D consistency into the 2D weights of the LLM. By training with $\mathcal{L}_{depth}$, our model learns to implicitly respect 3D boundaries (e.g., occlusion) using only 2D RGB inputs during inference.
\end{itemize}
Therefore, GASP does not replace RoPE but complements it: \textit{RoPE provides the "where" within the image grid, while GASP teaches the "what" and "which" across the spatio-temporal manifold.}

\end{document}